%% file: main.tex
\documentclass[12pt]{article}
\usepackage[T1]{fontenc}
\usepackage[utf8]{inputenc}
\usepackage{lmodern}
\usepackage[a4paper,margin=2.5cm]{geometry}
\usepackage{graphicx}
\usepackage{booktabs}
\usepackage[font=small]{caption}
\usepackage{placeins}
\usepackage{float}
\usepackage{amsmath,amssymb}
\usepackage[authoryear]{natbib}
\usepackage{xcolor}
\usepackage{hyperref}
\definecolor{ManuscriptLink}{HTML}{155CA2}
\hypersetup{colorlinks=true,linkcolor=ManuscriptLink,citecolor=ManuscriptLink,urlcolor=ManuscriptLink}
\newcommand{\norm}[1]{\left\lVert#1\right\rVert}
\title{\textbf{A Spiking Neural Model of the Mental Lexicon: Representations, Similarity Structure, and Associative Retrieval in Perception and Production}}
\author{Bernd J.~Kröger$^{1,2}$\\[1ex]
{\small $^{1}$Medical School, RWTH Aachen University, Aachen, Germany}\\
{\small $^{2}$Kröger Lab, Belgium, \url{www.speechtrainer.eu}}\\[2ex]}
\date{}
\begin{document}
\clubpenalty=3000
\widowpenalty=3000
\maketitle
\begingroup\footnotesize\noindent Preprint, 6 November 2026.\par\endgroup
\begin{abstract}
A neural account of lexical access requires an explicit distinction between the structure of lexical representations, the associations linking representational levels, and the activity generated during retrieval. This paper specifies a spiking neural implementation of the mental lexicon within the Neural Engineering Framework and Semantic Pointer Architecture. Conceptual and phonological form representations are constructed recursively from relational graphs, whereas lemma representations provide independent lexical identities. The implementation contains 613 lexical entries and supports two directed processing pathways: perception maps a phonological form to its associated concept via a lemma, whereas production maps a concept to its associated phonological form via a lemma. Each pathway combines two associative-memory variants, winner-take-all (WTA) selection (competition) and thresholded transfer, and uses 199,400 leaky integrate-and-fire neurons. We formalize within-level similarity, cross-level associative retrieval, and the interpretation of decoded activity. Simulations show successful retrieval of eight selected words from the lexicon in both processing directions, with retrieval evaluated against all 613 lexical entries. These simulations use complete phonological forms as input for perception and complete concepts as input for production, illustrating lexical retrieval through the corresponding processing pathways. This paper provides a transparent description of lexical representations and neural retrieval mechanisms, supported by accessible code, lexical data, and online simulation documentation.
\end{abstract}
\input{sections/01-introduction}
\input{sections/02-framework}
\input{sections/03-representations}
\input{sections/04-architecture}

\input{sections/05-methods}
\input{sections/simulation-corpus}
\input{sections/06-results}
\input{sections/07-discussion}
\input{sections/08-resources}
\input{sections/09-conclusion}
\section*{Acknowledgments}
The author thanks Katharina Stille for her contributions to the earlier development and investigation of the mental-lexicon model and for the research underlying the published work cited here. This manuscript builds on that scientific foundation while documenting the present implementation.

ChatGPT assisted with drafting and revising the manuscript; the author substantially developed and critically revised the text and takes full responsibility for its scientific content and final wording.

\input{references}
\appendix
\input{sections/appendix}
\end{document}

%% file: sections/01-introduction.tex
\section{Introduction}
Lexical processing links knowledge about a referent or intended message with a word identity and its phonological form. A computational account must specify both the representations available at these levels and the operations that transform one representation into another. These requirements become particularly important when a model is implemented using spiking neurons: a diagram of connected lexical levels does not by itself explain what the neural populations represent, how a mapping is realized, or what a plotted activation value measures.

Previous work has implemented lexical processing within the Neural Engineering Framework (NEF) and Semantic Pointer Architecture (SPA). \citet{stille2020} investigated the mental lexicon in long-term and working memory and simulated naming with semantic and phonological cues. Related work addressed production--perception interactions and internal monitoring \citep{kroeger2016}, while \citet{kroeger2023} placed the approach within a broader framework for speech production. These studies provide the scientific context for the present work. The present study examines the same basic lexical organization of concepts, lemmas, and phonological forms within more narrowly defined retrieval tasks. The restriction concerns the task setting and the surrounding task architecture, rather than a reduction of the lexical inventory or its three representational levels. The present implementation is not a reproduction of the complete task architecture or the published time courses of those models.

The contribution here is a formal and implementation-level account of the lexical core. Three complementary aspects organize this account. First, our approach links lexical graph representations to their vector encodings and to the representation of those vectors in populations of spiking neurons. This makes it possible to trace how linguistic relations contribute to representational similarity and how the resulting representations are realized in neural activity. Second, we distinguish similarity within a representational level from association between levels: the former describes overlap among representations, while the latter links corresponding concepts, lemmas, and phonological forms for lexical retrieval. Third, we examine activation patterns at the lexical levels and assess which lexical item is retrieved at the output of each processing stage. This relates the similarity structure of the representations to the outcome of lexical processing.

To develop these three aspects, the remainder of this paper proceeds from the theoretical framework to the construction of a 613-entry lexicon and its implementation in separate perception and production pathways. We then present the simulation methods and baseline results for eight selected words from the lexicon, followed by a discussion of their interpretation, the scope of the model, and directions for further investigation. The processing tasks begin with complete phonological form or conceptual vectors. Acoustic recognition, visual object recognition, and articulation are outside the present model. This restricted scope allows the relationships among representational geometry, associative retrieval, and neural realization to be examined directly. The transition from phonological form or syllabic representation to articulatory control is reserved for a separate account.

The primary aim of this paper is to provide a transparent description of the model, supported by illustrative simulations. The resulting simulation corpus provides a basis for broader evaluations, including additional inputs, model realizations, and comparisons with human behavior, as outlined in Section~\ref{sec:limitations}.

%% file: sections/02-framework.tex
\section{Theoretical and Computational Framework}
\subsection{Representational levels}
The model represents each entry in a unified lexicon at three complementary levels: the conceptual level, the lemma level, and the phonological level. For each lexical entry $i$, a conceptual vector $\mathbf c_i$, a lemma vector $\boldsymbol\ell_i$, and a phonological form vector $\mathbf p_i$ represent the corresponding concept, lemma, and phonological form of the same word. All vectors have dimensionality $D$. Each level uses independently generated base vectors to construct its representations within a level-specific coordinate system. The correspondence between the three representations of a lexical entry is specified through explicit cross-level associations. In the current implementation, the lemma level serves as a lexical identity interface; shared grammatical features, inflectional structure, and syntactic processing remain topics for future extensions.

Representations at the conceptual and phonological levels are structured through relations to other representations within the same level. For example, the phonological form of a word is related to representations of its phonological components; conceptual representations similarly incorporate associations with other conceptual representations. These relations are specified as graphs: nodes identify the represented concepts, phonological forms, or their components, and edges specify the relations between them. Each node is assigned a vector that incorporates its relations to other nodes, yielding a stored vector description of the lexical structure. These vectors are used to construct the neural network. During simulation, populations of spiking neurons encode time-dependent approximations to the vectors in their collective activity. The model thus connects two descriptions: a specification of lexical structure in terms of graphs and vectors, and its neural realization in terms of spiking population activity. Graph edges contribute to the construction of the stored vectors, while synaptic connections implement the transformations between neurally represented states.

\subsection{Neural representation and transformation}
The NEF describes how populations of neurons represent quantities and approximate transformations of those quantities \citep{eliasmith2003}. A useful rate-level description of the encoding of a vector $\mathbf x$ is
\begin{equation}
 a_j(\mathbf x)=G_j\!\left(\alpha_j\mathbf e_j^{\mathsf T}\mathbf x+J_j^{\mathrm{bias}}\right),
\end{equation}
where $\mathbf e_j$ is an encoder, $\alpha_j$ a gain, $J_j^{\mathrm{bias}}$ a bias current, and $G_j$ the neuronal response function. The simulations use spiking leaky integrate-and-fire (LIF) neurons, rather than substituting these rates for spike generation. A decoded state can be written
\begin{equation}
 \hat{\mathbf x}(t)=\sum_j\mathbf d_j\,(h*s_j)(t),
 \label{eq:decode}
\end{equation}
where $s_j$ is a spike train, $h$ a filtering kernel, and $\mathbf d_j$ a decoder. Decoders are computed to approximate the specified represented quantity or transformation over sampled evaluation points. This construction is distinct from learning lexical associations through experience.

The model is implemented in Nengo \citep{bekolay2014} with NengoSPA components. State vectors are distributed over one-dimensional ensembles, so Eq.~\eqref{eq:decode} applies componentwise within the corresponding ensemble array. The architecture specifies a functional computation; it does not provide a one-to-one anatomical assignment for each array.

\subsection{Semantic pointers and structured representations}
The SPA uses high-dimensional vectors to support both distributed representation and symbol-like operations \citep{eliasmith2013}. In the present model, three operations construct a structured vector from a node's base vector and its relations to other nodes: circular convolution, addition, and normalization. Circular convolution binds a vector specifying a relation to the vector representing its target. For example, binding the relation vector for \texttt{AssocWith} to a handle-related conceptual vector encodes the contribution ``associated with a handle.'' The resulting vector has the same dimensionality $D$ as its inputs and expresses the target in the context of that relation. Binding thus produces a new pattern within the same vector space, with a direction typically distinct from those of either input.

Addition combines these bound contributions with the node's own base vector. For example, the conceptual representation of \emph{Tasse} combines its base vector with contributions encoding associations with a dish-related concept and a handle-related concept. This sum forms a single vector containing contributions from several aspects of the word's conceptual structure. Normalization then rescales the sum to length 1 while preserving its direction. The stored representations therefore lie on the unit sphere in the $D$-dimensional space. A successfully retrieved neural state approximates one of these unit-length vectors; its decoded length can vary during activation buildup and through fluctuations in spiking activity.

Structured knowledge representation provides a broader context for combining relations and their targets in vector form \citep{crawford2016}. A central consequence in the present model is the emergence of representational similarity from shared structure. For example, \emph{Tasse} and \emph{Kanne} both include the same handle-related bound contribution. Such shared contributions can bring their normalized vectors closer together on the unit sphere. Groups of representations with overlapping structure can consequently occupy nearby regions of this sphere, with their precise positions determined by the full combination of base vectors and bound contributions. In this way, the graph structure shapes the geometry of conceptual and phonological similarity within each level.

In the present model, the lexical vectors are specified computationally and then realized in populations of spiking neurons. Their construction starts from a predefined lexical graph and randomly generated base and relation vectors, which are combined using the encoding operations described above. The resulting representations therefore reflect both the supplied lexical structure and the chosen encoding scheme. This dependence on the chosen lexical specification is a general property of the modeling approach and applies equally to alternative lexicons. Establishing how closely the resulting similarities correspond to human semantic or phonological similarity judgments requires an independent empirical comparison. This introduces a degree of arbitrariness in the assignment of vectors and constitutes a limitation of the current model: lexical representations and their associations are specified in advance rather than acquired through learning from linguistic experience. The graph constrains the relational structure, while the random vectors provide a reproducible numerical encoding of that structure. A fixed vector seed records one particular assignment within this scheme, and alternative seeds provide other numerical realizations of the same specified lexical structure, as explained in Section~\ref{sec:graph-encoding}. Vector construction precedes the neural realization, and the simulation examines how the resulting representations can be retrieved and mapped between lexical levels through spiking population activity. Learning these representations and associations from experience is a direction for future development.

%% file: sections/03-representations.tex
\section{Lexicon Construction and Representational Similarity}
\subsection{Lexical inventory and provenance}
The active dataset contains 613 lexical entries. Each entry identifies a word label and its corresponding concept, lemma, and phonological form nodes. The conceptual graph contains 1,205 nodes, including 593 nodes classified as deep-layer nodes in the source data; the phonological graph contains 1,616 nodes, including 1,003 deep-layer nodes. Deep-layer nodes supply representational structure rather than additional retrieval candidates; the underlying representational organization is described by \citet{stille2020}. One conceptual lexical root also belongs to the deep-layer category, so the conceptual counts are not disjoint categories that can simply be added.

The graphs were extracted from the existing WWT source data \citep{stille2020}. The current dataset documents 13 restored phonological mappings and one component correction for the entry \emph{Verbringen}. Original and restored links are recorded explicitly. These repairs are changes introduced in the current dataset relative to the source files used by Katharina Stille in the earlier modeling work \citep{stille2020}.

In this paper, eight words are used for detailed display and as baseline inputs: \emph{Tasse}, \emph{Tasche}, \emph{Kanne}, \emph{Katze}, \emph{Teller}, \emph{Topf}, \emph{Stuhl}, and \emph{Schrank}. This selection keeps the user interface and the detailed presentation manageable while providing a compact set of examples to illustrate how the model operates. The underlying model uses the full lexicon of 613 entries, all of which remain available as retrieval candidates in the associative memories.

\subsection{Recursive graph encoding}
\label{sec:graph-encoding}
Let $E(x)$ be the outgoing relation--target pairs of node $x$, $\mathbf b_x$ its random base vector, and $\mathbf r_\rho$ the vector assigned to relation $\rho$. A node is encoded as
\begin{align}
 \mathbf u_x &= \mathbf b_x+\sum_{(\rho,y)\in E(x)}\mathbf r_\rho\circledast\mathbf v_y,\\
 \mathbf v_x &= \frac{\mathbf u_x}{\norm{\mathbf u_x}}.
 \label{eq:recursive}
\end{align}
For indices $k=0,\ldots,D-1$, circular convolution is
\begin{equation}
 (\mathbf a\circledast\mathbf b)_k=\sum_{m=0}^{D-1}a_m b_{(k-m)\bmod D}.
\end{equation}
A leaf has no outgoing relations and is represented by its normalized base vector. Non-leaf nodes incorporate the recursively constructed target vectors, not separate target identity vectors. Each node is normalized after summation. The implementation checks for directed cycles because this recursive procedure requires an acyclic dependency structure.

Base and relation vectors are obtained by normalizing independent Gaussian draws. They are ordinary random unit vectors; no unitary-convolution constraint is imposed. A \emph{vector seed} fixes the pseudorandom choices from which the vectors are generated. In this implementation, each vector's random stream is derived deterministically from the seed, its representational level and role, and its node or relation identifier. The seed therefore specifies one numerical realization of the vector vocabulary, including the independently generated lemma vectors.

This choice makes the arbitrariness of the numerical encoding explicit. The lexical graph specifies which nodes and relations are present and which components are shared; the seed selects concrete vector values within that specification. For example, different seeds assign different numerical vectors to the handle-related component shared by \emph{Tasse} and \emph{Kanne}, while both words continue to share that component. The graph-defined relations are preserved across these realizations, and the exact similarities between word vectors can vary. The resulting geometry reflects both the specified lexical structure and its particular numerical realization.

Holding the graph, vector dimensionality, encoding procedure, identifiers, and software settings fixed, the same seed reproduces the same vector vocabulary. Fixing the seed thus preserves one chosen configuration for repeated inspection or comparisons between model conditions. Deliberately changing it generates alternative configurations of the same lexical specification; recording each alternative seed makes those configurations reproducible as well. This provides a way to examine how retrieval depends on the numerical encoding. The role of the seed is to control that choice; the linguistic adequacy of the resulting similarities remains a question for independent empirical comparison.

Because the random streams are tied to vector identifiers, existing vectors also remain stable when unrelated entries are added, provided that their dependency graphs remain unchanged. This supports comparisons between the eight-entry fixture and the 613-entry dataset with identical vectors for the shared entries. The previous WWT vectors used in the model developed by Katharina Stille were generated with a different randomization procedure.

\subsection{Conceptual similarity}
Concept nodes use the relation \texttt{AssocWith}. For example, the source graph associates \emph{Tasse} with both a dish-related node and a handle-related node, whereas \emph{Kanne} includes the handle-related node. Their encodings therefore contain a common bound contribution. The resulting overlap depends on all contributions and the normalization in Eq.~\eqref{eq:recursive}; it is not assigned directly as a fixed pairwise similarity value.

Shared components make similarity interpretable in terms of the conceptual relations encoded in the graph. Inspecting these relations helps explain why particular word vectors overlap and which aspects of their conceptual structure contribute to that overlap. For example, the source concept for \emph{Tasche} includes a connection to a vending-machine-related visual node, which contributes to its vector through the same encoding rule. The resulting similarities thus provide a way to examine how the associations specified in the lexical dataset shape the geometry of the conceptual representations.

\subsection{Phonological similarity}
Phonological form nodes use the relation \texttt{InclPhon} to incorporate recursively represented components. For example, \emph{Tasse} and \emph{Tasche} share the source component \texttt{P\_St\_ta}, while \emph{Kanne} and \emph{Katze} share \texttt{P\_St\_ka}. Common components at deeper levels can also contribute to overlap. Phonological similarity in this model thus expresses the overlap between vector representations constructed from the components and relations specified in the phonological graph. Shared structure contributes to this overlap through the recursive encoding procedure, with the resulting similarity determined by the complete composition and normalization of the word vectors. The measure therefore provides an explicit link between the phonological structure represented in the lexicon and the similarity values used to interpret lexical retrieval.

To construct a phonological representation, the model first binds each component vector to the vector for its relation, such as \texttt{InclPhon}, and then adds these contributions to the node's base vector. For example, if a node includes two components with bound contributions $\mathbf q_1$ and $\mathbf q_2$, their combined contribution is $\mathbf q_1+\mathbf q_2$. Addition gives the same result when these contributions are processed in the reverse order. The resulting representation therefore depends on the components and relations specified in the graph and remains stable across different orderings of the same list of graph connections. Position within a word---for example, whether a component belongs to the first or second syllable---is a separate aspect of phonological structure. Encoding such a distinction requires it to be expressed in the component identities or relations supplied to the encoding procedure. The graph thus specifies the phonological content and structural distinctions that are combined into a static vector representation for lexical retrieval. These structural distinctions describe positions within a word; they do not specify when its components become active or how long they remain active during articulation. The present encoding therefore does not generate a temporal sequence of syllables or articulatory gestures.

\subsection{Within-level similarity and cross-level association}
Within-level similarity describes the overlap between lexical representations on the same representational level. It applies to all three levels: conceptual, lemma, and phonological. In the following notation, $L$ denotes any one of these levels. For normalized vectors representing lexical entries $i$ and $j$ at the selected level, define
\begin{equation}
 S^{(L)}_{ij}=(\mathbf v_i^{(L)})^{\mathsf T}\mathbf v_j^{(L)}.
 \label{eq:similarity}
\end{equation}
This quantity is both a dot product and a cosine similarity. Applied separately to the conceptual vectors $\mathbf c_i$, lemma vectors $\boldsymbol\ell_i$, and phonological form vectors $\mathbf p_i$, it yields three similarity matrices, each describing the static representational geometry of its level. Values near zero indicate small overlap. Negative values express opposing vector components in this geometric measure; their interpretation in terms of linguistic relations or neural activity requires additional evidence. Independent random unit vectors have expected dot product zero and variance $1/D$, providing a reference for incidental overlap. The distribution of similarities among structured vectors additionally reflects their encoded relations.

At the conceptual and phonological levels, the graph-based construction introduces shared structure into the vectors. Conceptual overlap can reflect common associations, while phonological overlap can reflect common components of word forms. The lemma level currently represents lexical identity using independently generated vectors. Similarity is measured there by the same equation, with the resulting overlap arising from the random-vector construction. These lemma vectors are approximately orthogonal, providing distinct identities for the mappings between concepts and phonological forms.

A possible extension would enrich lemma representations with shared grammatical features. For example, lemmas belonging to the same word class, such as nouns or verbs, could include a common contribution encoding that class alongside their individual identity contributions. Other grammatical properties could be incorporated in the same way. Such a construction would introduce systematic lemma-level similarities based on grammatical structure. This extension would require an explicit specification of the features and their encoding; the present implementation uses the independent identity vectors described above.

Cross-level association links the representations of the same lexical entry across levels. The model specifies both concept--lemma associations through the pairs $(\mathbf c_i,\boldsymbol\ell_i)$ and lemma--phonological form associations through the pairs $(\boldsymbol\ell_i,\mathbf p_i)$. In production, these associations support retrieval from concept to lemma and from lemma to phonological form; in perception, they support retrieval in the reverse direction. Their correspondence is established by the shared lexical entry index $i$. Thus, within-level similarity describes the geometry among representations on one level, while cross-level association specifies which representation on another level belongs to the same lexical entry. The associative memories implement these correspondences between the independently constructed vector spaces.

%% file: sections/04-architecture.tex
\section{Network Architecture and Associative Retrieval}
\label{sec:architecture}
\subsection{Directed processing pathways}
\begin{figure}[htbp]
\centering
\includegraphics[width=\linewidth]{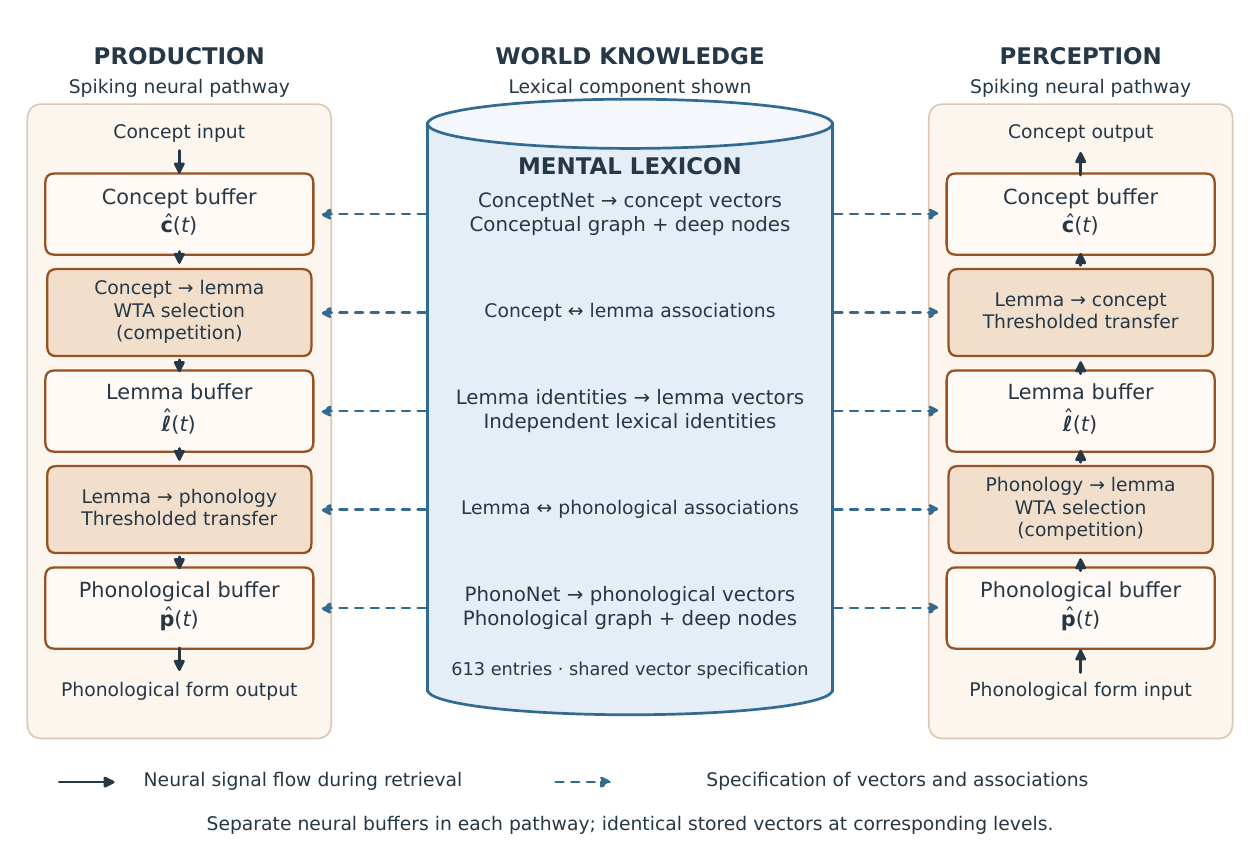}
\caption{Stored lexical knowledge and its neural realization in the present model. The central cylinder denotes the graphs, vectors, and cross-level associations specifying the mental lexicon within world knowledge. Left: production proceeds from concept through lemma to phonological form. Right: perception proceeds from phonological form through lemma to concept. Each pathway contains separate spiking state buffers and two neural associative memories, one of each variant: WTA selection (competition) and thresholded transfer. Both variants apply thresholding; WTA selection additionally implements competition through lateral inhibition. Corresponding buffers on the two sides use identical stored vectors. Solid arrows indicate neural signal flow; dashed arrows indicate how the shared lexical specification defines the buffers' vocabularies and the memories' associations during network construction. The cylinder is a representation of stored knowledge, not an additional spiking module supplying signals during simulation. WTA: winner-take-all. Original diagram of the lexical processing architecture, informed by \citet[Figure~1]{stille2020}.}
\label{fig:architecture}
\end{figure}

Figure~\ref{fig:architecture} distinguishes the stored lexical knowledge from its use in neural processing. Its organization follows the distinction between a central knowledge repository and lateral processing pathways illustrated in Figure~1 of \citet{stille2020}, here redrawn for the present lexical core. The central cylinder represents the mental lexicon as the lexical component of world knowledge. It contains the conceptual and phonological graphs, their derived vectors, the lemma identity vectors, and the associations linking the three representations of each lexical entry. ConceptNet and PhonoNet denote the conceptual and phonological graph structures, including their deep-layer components.

The left and right columns show the neural realization of lexical processing. Production maps a concept to its associated lemma and then to its phonological form; perception maps a phonological form to its lemma and then to its concept. Each column contains three neural state arrays, referred to here as buffers, and two associative memories. The buffers represent the currently active vectors through the collective activity of spiking neurons \citep{eliasmith2013}. The associative memories implement the mappings between levels using the entry correspondences specified in the central repository.

Production and perception are implemented as separate networks, each with its own buffers and associative-memory populations. Corresponding buffers nevertheless use the same vector vocabulary: a given lexical entry has the same stored conceptual vector in both pathways, the same lemma vector in both pathways, and the same phonological form vector in both pathways. These shared vectors are reproduced from the same lexical data and vector seed. The time-dependent neural states are specific to each pathway and its input. Thus, the two columns represent separate neural realizations of a shared lexical specification, rather than two simultaneous directions through one set of buffers.

The dashed links in Figure~\ref{fig:architecture} express this dependence on the lexical specification. During network construction, the stored vectors define the state vocabularies and the input and output vectors of the associative memories. During simulation, retrieval proceeds through the neural connections indicated by solid arrows. Both transformations in each pathway are implemented by heteroassociative memories, which retrieve a stored output vector associated with an input vector from another lexical level. The labels WTA selection (competition) and thresholded transfer in Figure~\ref{fig:architecture} identify two variants of this common associative-memory mechanism. The first memory performs WTA selection (competition) on the way to the lemma buffer; the second performs thresholded transfer from the lemma buffer to the target level. The state self-feedback gain is zero, so the buffers follow their incoming signals, and interlevel processing follows the indicated feedforward direction. Section~\ref{sec:neural-implementation} explains this parameter and its role in the present architecture.

\subsection{Formal specification of associative memories}
Each associative memory links two adjacent lexical levels. Its input vocabulary contains the stored vectors against which the incoming neural state is compared; its output vocabulary contains the vectors that can be retrieved through the corresponding associations. In production, the first memory uses conceptual vectors as its input vocabulary and lemma vectors as its output vocabulary. The second uses lemma vectors as input and phonological form vectors as output. In perception, the first memory maps phonological form vectors to lemma vectors, and the second maps lemma vectors to conceptual vectors. The terms input and output vocabulary therefore refer to the role of a representational level in a particular memory.

For any one of these four mappings, let the columns of $U\in\mathbb R^{D\times N}$ contain the stored input vectors and the columns of $V\in\mathbb R^{D\times N}$ contain their associated output vectors, with $N=613$. Column $i$ in both matrices refers to the same lexical entry. For example, in the concept-to-lemma memory, the column containing the conceptual vector for \emph{Kanne} in $U$ is paired with the column containing its lemma vector in $V$. In the implementation, this correspondence is established through a common entry identifier, or key, in the two vocabularies. NengoSPA's \texttt{by-key} mapping uses these identifiers to assemble the input and output vectors in matching column order. The pairing thus follows lexical identity and remains valid when the vocabularies themselves list their entries in different orders.

Retrieval comprises three operations: comparison with the stored input vectors, selection among the resulting matches, and construction of the associated output. Their functional organization is
\begin{equation}
 \mathbf q(t)=U^{\mathsf T}\mathbf x(t),\qquad
 \mathbf z(t)=\mathcal F[\mathbf q](t),\qquad
 \mathbf y(t)=V\mathbf z(t).
 \label{eq:am}
\end{equation}
Here $\mathbf x(t)$ is the $D$-dimensional vector supplied by the preceding neural buffer. The first operation produces $N$ match values: each $q_i(t)$ is the dot product of the incoming vector with the stored input vector for lexical entry $i$. The neural selection mechanism $\mathcal F$ transforms these matches into the $N$ decoded selection outputs $z_i(t)$. The final operation combines the stored output vectors using these selection outputs as weights, yielding the $D$-dimensional vector $\mathbf y(t)$ for the next buffer. When the selection output for \emph{Kanne} dominates, for example, the retrieved vector is dominated by the representation associated with \emph{Kanne} at the output level.

Figure~\ref{fig:architecture} identifies the two associative-memory variants as \emph{WTA selection (competition)} and \emph{thresholded transfer}. WTA selection (competition) denotes the competitive associative memory used at the first transition in each pathway: concept to lemma in production and phonological form to lemma in perception. Thresholded transfer denotes the associative memory used at the second transition: lemma to phonological form in production and lemma to concept in perception. Both variants use a threshold to admit sufficiently strong matches. The WTA selection variant additionally introduces competition between candidates to select a lemma identity; the thresholded transfer variant passes qualifying lemma matches to their associated target representations. WTA selection performs the cleanup role \citep{stewart2011}: it favors a stored lexical candidate when the input also matches other candidates. Here, cleanup is combined with a mapping to another lexical level, giving a heteroassociative memory. The matrices $U$ and $V$ specify the stored correspondences, and $\mathcal F$ supplies the nonlinear thresholding and selection behavior. Section~4.3 describes the two selection mechanisms in detail.

Equation~\eqref{eq:am} separates these functional operations to make their roles explicit. In the spiking network, signals also undergo temporal filtering at synaptic connections. These connection filters are low-pass filters that shape how neural signals evolve over time \citep{eliasmith2003}; cleanup memories instead operate on the match values to select stored representations. The compact equation leaves the synaptic filter kernels implicit. The resulting time courses depend jointly on these filters, neuronal dynamics, and the selection circuit, with the principal time constants specified in Section~4.4.

\subsection{WTA selection (competition) and thresholded transfer}
The two associative-memory variants shown in Figure~\ref{fig:architecture}, \emph{WTA selection (competition)} and \emph{thresholded transfer}, share the same input-comparison and output-mapping organization. Here, \emph{thresholding} denotes the operation that admits matches above a threshold and is used in both variants. The term \emph{competition} denotes the additional interaction between candidates in WTA selection, implemented through lateral inhibition. Thresholded transfer evaluates each candidate against the threshold independently; WTA selection additionally allows strongly activated candidates to suppress competing responses. In NengoSPA, WTA selection (competition) is implemented by \texttt{WTAAssocMem}, and thresholded transfer by \texttt{ThresholdingAssocMem} \citep{nengospaAM}. These are two variants of associative memory, distinguished by their selection mechanisms.

Both memories use a match threshold of $\theta=0.3$. For the thresholded transfer variant used in the second memory, an idealized static target function is
\begin{equation}
 z_i=\begin{cases}q_i,&q_i>\theta,\\0,&q_i\leq\theta.\end{cases}
 \label{eq:threshold}
\end{equation}
This target function preserves the amplitude of each match that exceeds the threshold. The spiking implementation approximates this behavior with neural dynamics and decoding variability. Several entries can pass when several matches exceed threshold, so this variant supports the transfer of multiple qualifying associations.

The WTA selection (competition) variant used in the first memory combines lateral inhibition with a constant decoded output function of one for a surviving selection. In an ideal settled state, it approximates a single selected output vector rather than preserving the winning input's amplitude. The implementation inhibits each selection population using the other populations' raw thresholded outputs. The inhibition scale is 1.0 and the inhibitory synapse has a 5-ms time constant. Selection transients and ties cannot be described fully by an instantaneous maximum operation.

This arrangement retains the division of roles in the preliminary WWT source implementation developed by Katharina Stille: WTA selection (competition) provides cleanup on the way to the lemma level, followed by thresholded transfer to the target level. The first memory receives a structured conceptual or phonological representation whose shared components can yield matches to several lexical candidates. WTA selection (competition) supports the selection of a distinct lemma identity from these matches. The second memory then retrieves the phonological form or concept associated with that selected lemma. Because lemma vectors encode distinct lexical identities and are approximately orthogonal, thresholded transfer is used at this stage to pass sufficiently strong matches to their associated outputs while preserving their amplitudes.

These distinctions give the activation curves a specific computational interpretation: the WTA selection (competition) memory provides a competitive selection signal, the thresholded transfer memory provides an amplitude-sensitive signal, and the state arrays represent distributed vectors. For example, an entry output near 1 in the WTA selection (competition) memory reflects the configured signal amplitude for a selected candidate. In the thresholded transfer memory, several entries can contribute simultaneously, with each entry's output specifying its weight in the retrieved vector. The curves thus describe selection and signal strength over time. Section~\ref{sec:neural-probabilistic} discusses how this mechanistic account relates to models that explicitly represent uncertainty over lexical alternatives.

\subsection{Neural implementation and parameters}
\label{sec:neural-implementation}
The neural buffers in Figure~\ref{fig:architecture} are implemented as \emph{state arrays}. An \emph{ensemble} is a group of neurons whose collective activity represents a quantity \citep{eliasmith2003}; an \emph{array} is a collection of such ensembles working together. The term \emph{state} refers to the information currently represented by their activity. In the present implementation, each ensemble represents one component of the buffer's current vector, and the ensembles together provide the complete neural representation of the buffer state.

The \emph{state self-feedback gain} specifies the factor by which a buffer's decoded output is scaled when fed back to its own input. Here, the gain is zero: each buffer is driven by the external input or the preceding associative memory. This choice matches the focus on input-driven lexical retrieval. In the perception and production tasks, the input representation is supplied continuously during retrieval and evaluation (Section~\ref{sec:tasks}), sustaining the signals passed through the successive processing stages. The sustained input thus provides the temporal support needed for the present tasks, while the buffers represent and transmit the available signals. After the driving signal subsides, residual activity decays according to the neuronal and synaptic dynamics. The zero gain applies specifically to buffer self-feedback; competition within the WTA associative memories continues to use lateral inhibition.

A possible next extension would give the input buffer a short-term maintenance function through controlled recurrent self-feedback. In perception, a briefly supplied phonological form vector would be retained in the phonological input buffer; in production, the corresponding role would be taken by the conceptual input buffer. The retained activity would continue to drive associative retrieval after the external cue has ended. This would separate the brief presentation of a representation from its internal maintenance and subsequent use, providing a closer match to tasks in which information is available only temporarily. This recurrent input buffer is proposed as a further development; the reported simulations use sustained input during retrieval. Section~\ref{sec:limitations} outlines the corresponding implementation and evaluation steps.

With $D=512$ one-dimensional ensembles and 50 LIF neurons per ensemble, each state array contains $512\times50=25{,}600$ neurons. Each associative memory uses 100 neurons per lexical candidate, giving $613\times100=61{,}300$ neurons. The three buffers and two associative memories in each pathway therefore contain a total of
\begin{equation}
 3(25{,}600)+2(61{,}300)=199{,}400.
\end{equation}
Input-to-state and intermodule connections use low-pass synapses with a 10-ms time constant, which shape the transmission of activity between neural populations. Together with the 5-ms inhibitory time constant of the WTA selection (competition) memory, these synaptic parameters contribute to the dynamics of the neural architecture.

Two separate seeds control the numerical representation and its neural realization. The \emph{vector seed}, fixed at 10, selects the vector vocabulary described in Section~\ref{sec:graph-encoding}. Once that vocabulary has been constructed, the \emph{network seed}, fixed at 11, controls sampled properties of the neural populations, including encoders and neuronal response parameters. Keeping the vectors fixed while changing the network seed produces an alternative neural realization of the same representational geometry and intended associations. Changing the vector seed instead supplies an alternative vocabulary, for which the neural mappings are then constructed. The two seeds therefore make it possible to investigate variation in the numerical encoding separately from variation in the neural implementation. With fixed lexical data, model settings, and software, retaining both seeds reproduces the chosen realization used throughout the reported baseline simulations.

Table~\ref{tab:parameters} summarizes the implemented settings, including the representation sizes, neuron counts, feedback, match threshold, and synaptic time constants. The perception and production tasks applied to this architecture, including trial timing, recording, and evaluation, are specified separately in Section~\ref{sec:tasks}.

\begin{table}[htbp]
\centering\small
\begin{tabular}{ll}\toprule
Quantity & Value\\\midrule
Lexical candidates / vector dimensions & 613 / 512\\
Vector seed / network seed & 10 / 11\\
Dimensions per state ensemble & 1\\
Neurons per state ensemble / memory entry & 50 / 100\\
Neurons per directed pathway & 199,400 LIF\\
State self-feedback gain & 0\\
Associative-memory match threshold & 0.3\\
Intermodule / WTA inhibitory time constant & 10 / 5 ms\\\bottomrule
\end{tabular}
\caption{Representation and neural architecture parameters of the 613-entry model. Library defaults not overridden by the model are preserved by the pinned software environment.}\label{tab:parameters}
\end{table}
\FloatBarrier

%% file: sections/05-methods.tex
\section{Task Realization Using the Neural Architecture}
\label{sec:tasks}
The architecture described in Section~\ref{sec:architecture} provides the neural buffers and associative memories for two lexical retrieval tasks: perception and production. Task realization specifies which representation is supplied as input, when it is presented, which output is recorded, and how retrieval success is assessed. Both tasks use the eight selected words introduced in Section~3.1 and the complete 613-entry lexical inventory.

\subsection{Inputs and trial procedure}
\label{sec:task-procedure}
In the \emph{perception task}, the complete phonological form vector of a selected word is supplied to the phonological buffer, and the pathway retrieves its associated concept through the lemma level. In the \emph{production task}, the complete conceptual vector is supplied to the concept buffer, and the pathway retrieves its associated phonological form through the lemma level. These tasks assess the transfer of already specified lexical representations. They do not include recognition of an acoustic waveform or inference of a concept from an image.

Each trial lasts 1.1~s (1,100~ms). The simulator advances the neural dynamics in steps of 1~ms; this simulation step specifies the numerical time increment. The input interval extends from 150~ms to immediately before 750~ms after trial onset. During this interval, the selected word's input vector is supplied continuously; the external input is zero before and after it. The trial therefore includes an initial period, a period of sustained input, and a period following input offset.

For each processing direction, one simulator is constructed, and its neural and filter state is reset between words. The input pointer changes from trial to trial, while the stored vectors, neural parameters, and associative mappings remain fixed. The vector and network seeds are those specified in Table~\ref{tab:parameters}. These are deterministic simulation runs for a fixed realization, not independent samples of a participant population.

\subsection{Recording and retrieval criterion}
\label{sec:task-evaluation}
State and selection probes record the resulting signals through an additional low-pass filter with a 20-ms time constant. This observation filter provides moderate temporal smoothing, reducing rapid fluctuations and making the course of lexical activation easier to inspect and compare. It acts on the recorded readout, while neuronal activity and synaptic interactions within the network remain unchanged. The curves thus provide a smoothed view of the ongoing network computation. Brief transients and the apparent onset and decay of recorded activity reflect this additional filtering as well as the network's own dynamics.

At each lexical level, the stored vocabulary vectors provide references against which the recorded neural state can be evaluated. For a recorded decoded state $\hat{\mathbf x}^{(L)}(t)$, the score for lexical entry $i$ is the dot product
\begin{equation}
 A_i^{(L)}(t)=(\mathbf v_i^{(L)})^{\mathsf T}\hat{\mathbf x}^{(L)}(t),
 \label{eq:activation}
\end{equation}
where $\mathbf v_i^{(L)}$ is the stored vector for that entry at level $L$. The same readout is used at the conceptual, lemma, and phonological levels. These scores form the time-dependent similarity curves for the state buffers. The associative memories' decoded selection outputs $z_i(t)$ are recorded separately, as specified in Eq.~\eqref{eq:am}.

The evaluation interval is the portion of the recorded trial used to assess retrieval. For the final state of each pathway, scores are averaged over $0.45\leq t<0.70$~s (450--700~ms) and evaluated against all 613 candidates. This window falls within the sustained-input period. A response is accepted if the largest score exceeds 0.3 and exceeds the second-largest score by more than 0.1. It is correct if the accepted candidate is the input word's associated lexical entry. These evaluation criteria determine whether the recorded response is accepted; the associative-memory match threshold in Table~\ref{tab:parameters} determines how the neural circuit processes candidate matches. Their shared numerical value of 0.3 serves these distinct purposes. Table~\ref{tab:task-settings} summarizes the trial, recording, and evaluation settings shared by the perception and production tasks.

\begin{table}[htbp]
\centering\small
\begin{tabular}{ll}\toprule
Quantity & Value\\\midrule
Simulation time step & 1 ms\\
Trial duration & 1,100 ms\\
Input interval & $[150,750)$ ms\\
Observation filter time constant & 20 ms\\
Evaluation interval & $[450,700)$ ms\\
Acceptance score threshold & 0.3 (strictly exceeded)\\
Acceptance margin over second-ranked candidate & 0.1 (strictly exceeded)\\\bottomrule
\end{tabular}
\caption{Settings for realizing and evaluating both lexical retrieval tasks using the neural architecture. Intervals are measured from trial onset; $[a,b)$ includes the start time $a$ and ends immediately before $b$.}\label{tab:task-settings}
\end{table}

\FloatBarrier

%% file: sections/simulation-corpus.tex
\section{Simulation Corpus}
\label{sec:simulation-corpus}
The \emph{simulation corpus} comprises eight perception trials and eight production trials: one for each selected word in each processing direction, giving sixteen trials in total. All trials use the complete 613-entry lexical inventory, the procedure in Section~\ref{sec:tasks}, and the fixed model configuration in Table~\ref{tab:parameters}. Each recorded result is associated with its input word, processing direction, model settings, and random seeds, and identifies the retrieved entry, target score, and margin over the next-ranked candidate. Appendix~\ref{app:implementation} identifies the numerical reports and recordings.

These trials constitute the \emph{baseline}, meaning the behavior of the intact network with all neurons available. The implementation also supports simulations of impaired networks through \emph{ablation}, which silences selected neurons in buffers or associative memories (see Section~\ref{sec:limitations}).

%% file: sections/06-results.tex
\section{Results}
\label{sec:results}
The resulting model comprises 613 lexical entries and supports retrieval in both the production and perception directions; under intact conditions, it correctly retrieves all eight entries tested in the present simulation corpus, with all 613 entries participating as candidates.

\subsection{Representational structure}
\begin{figure}[htbp]
\centering\includegraphics[width=\textwidth]{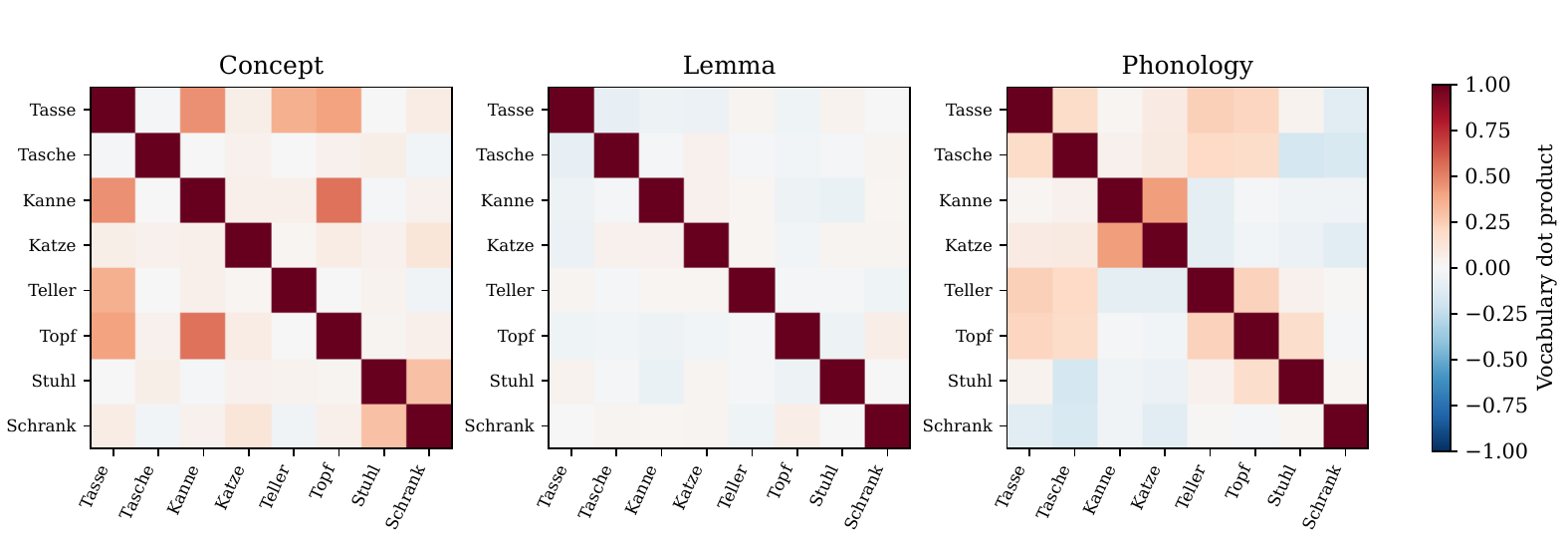}
\caption{Within-level dot products for the eight selected words, extracted from the 613-entry vocabulary ($D=512$, vector seed 10). All vocabulary vectors are normalized. The shared color scale spans $-1$ to $1$. The matrices describe static geometry, not neural activity or retrieval probabilities.}\label{fig:geometry}
\end{figure}

The model represents similarity relationships between lexical entries at both the conceptual and phonological levels. These relationships can be examined by comparing the corresponding vectors, providing an overview of the structure within the lexical inventory. Figure~\ref{fig:geometry} illustrates this structure for the eight selected entries by displaying their pairwise similarities at each of the three lexical levels.

At the conceptual level, for example, \emph{Kanne}, \emph{Tasse}, and \emph{Topf} share a handle-related component in the lexical graph. This common component contributes to the overlap between their conceptual vectors. At the phonological level, shared form components likewise contribute to similarity, as illustrated by \emph{Kanne} and \emph{Katze}, which share the component \texttt{P\_St\_ka}. The lemma vectors, in contrast, provide independently generated lexical identities. The matrices thus make the similarity structure of the constructed representations visible before neural simulation; they do not measure human similarity judgments.

\subsection{Retrieval in both directions}
In the intact model, with no neurons ablated, all eight selected words are correctly retrieved in both directions under the stated full-lexicon criterion (Table~\ref{tab:baseline}). Production target scores range from 0.9464 to 0.9729; perception target scores range from 0.9481 to 0.9792. The smallest acceptance margin is 0.4110 in production and 0.4372 in perception. The result establishes successful transfer for these inputs in the archived realization. It is not an estimate of accuracy over the full lexical inventory.

\begin{table}[htbp]
\centering\small
\input{tables/baseline}
\caption{Baseline retrieval results for the sixteen trials in the simulation corpus, evaluated against all 613 candidates. Scores are mean final-state dot products over 0.45--0.70~s; margins are the differences between the largest and second-largest scores. Every listed trial retrieves the intended entry.}\label{tab:baseline}
\end{table}
\FloatBarrier

\subsection{Activation dynamics in the two pathways}
\begin{figure}[p]
\centering\includegraphics[width=\textwidth]{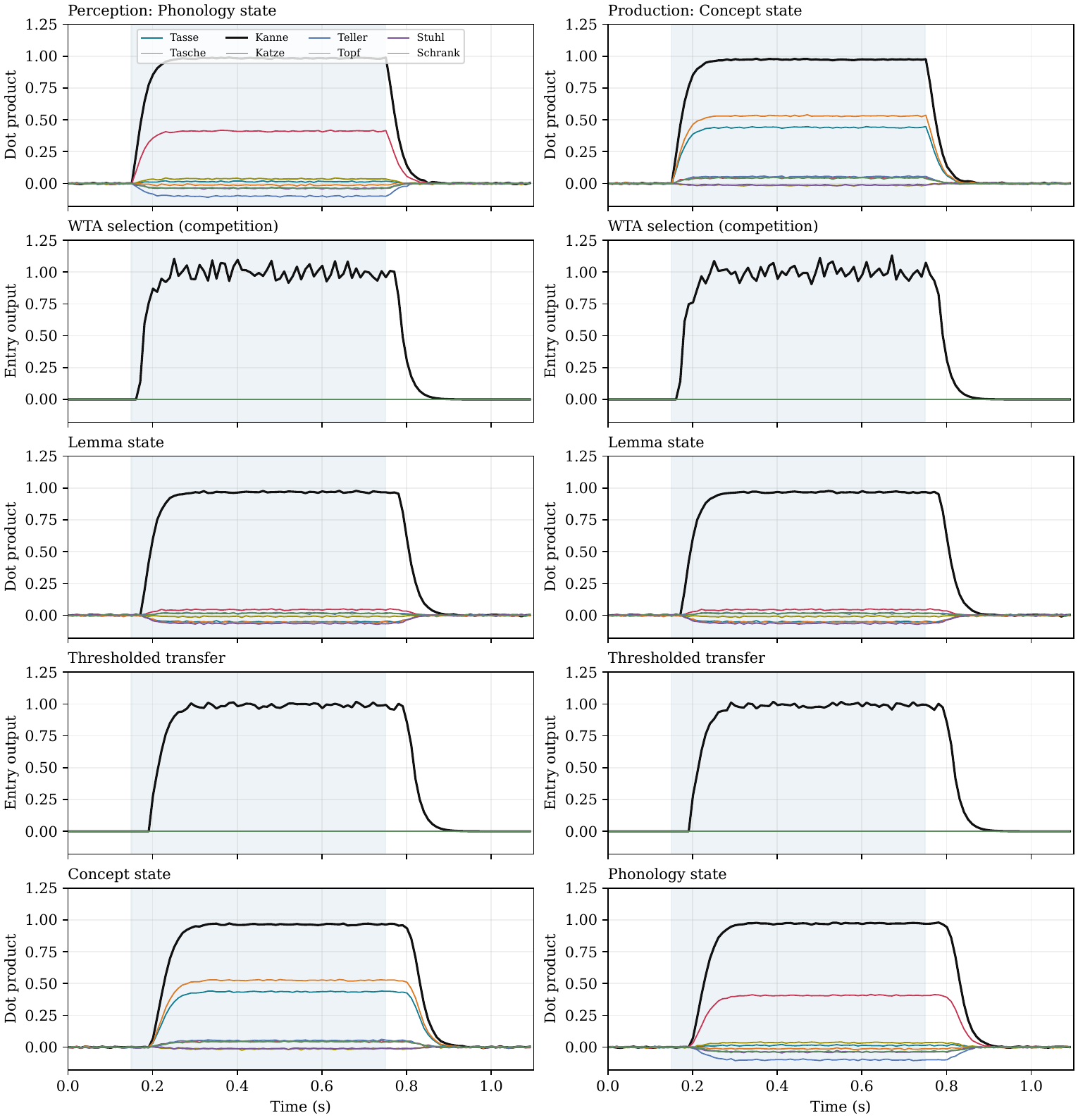}
\caption{Stored \emph{Kanne} simulations with 613 candidate entries. Left: perception; right: production. From top to bottom, each column shows the input-level state, WTA selection (competition) memory output, lemma state, thresholded transfer memory output, and final state. State panels show dot products with reference vectors; memory panels show decoded entry outputs. Shading marks the complete-vector input interval. The same eight entries are plotted in every panel, with the same color assignment in both columns; the legend in the upper-left panel applies throughout. In particular, the red curve represents \emph{Katze} in both perception and production. All 613 candidates participate in retrieval, but \emph{Krone}, the highest-scoring non-target candidate in the final production state, is not plotted; its score reported in the text comes from the full-lexicon numerical evaluation. The plotted export is rounded and sampled at approximately 10-ms intervals; its underlying simulation step is 1~ms.}\label{fig:traces}
\end{figure}

Figure~\ref{fig:traces} shows the stored \emph{Kanne} simulation traces for the perception and production tasks. Each column follows the processing direction through three state buffers and two associative memories. The state curves are the dot products defined in Eq.~\eqref{eq:activation}; the memory curves show decoded selection outputs. Shading identifies the interval during which the complete input vector is supplied.

In the production benchmark, the highest-scoring non-target final-state candidate for \emph{Kanne} is \emph{Krone}, with a score of 0.5619, while the target score is 0.9729. \emph{Krone} is not among the eight words plotted in Figure~\ref{fig:traces}; its score comes from the numerical evaluation over the full lexicon. The red curve in both columns represents \emph{Katze}, as specified by the shared color legend in the upper-left panel. This example also shows why displaying only eight words would be insufficient for deciding retrieval success: the nearest output competitor can be outside the set of selected words. The reported decision includes all 613 entries.

\FloatBarrier

\subsection{Interpreting activation patterns at the three lexical levels}
\label{sec:activation-interpretation}
The curves in Figure~\ref{fig:traces} now provide a concrete basis for interpreting activity at the conceptual, lemma, and phonological levels. Each state curve shows how the decoded neural representation projects onto the stored vector of a particular lexical entry over time. It is a readout of the distributed population activity, while the memory panels report the outputs of the associative selection circuits. Reading these panels together helps distinguish the representation currently carried by a buffer from the lexical associations selected during retrieval.

Representational overlap helps explain the patterns in the state panels. Because related lexical items share representational components, retrieving one item can produce positive scores for other items. For example, the conceptual representations of \emph{Kanne} and \emph{Tasse} share a handle-related component. A state representing only \emph{Kanne} can therefore yield a positive score for \emph{Tasse}. This score expresses overlap with the \emph{Tasse} vector; evidence that the \emph{Tasse} association has also been selected would come from the corresponding memory output. The same distinction between state projection and associative selection applies at all three lexical levels in both processing directions.

The dot-product score in Eq.~\eqref{eq:activation} carries information about both signal amplitude and alignment with a stored vector. The decoded state varies in length over time, so this readout differs from the cosine similarity between unit-length vocabulary vectors in Eq.~\eqref{eq:similarity}. A score can exceed one; a negative score indicates a projection in the opposite direction to the reference vector and does not by itself identify an inhibitory neural population.

The contribution of representational overlap can be stated mathematically. Suppose a state contains only a scaled target vector, $\hat{\mathbf x}(t)=a(t)\mathbf v_k$. Then
\begin{equation}
 A_i(t)=a(t)S_{ik}.
 \label{eq:overlap}
\end{equation}
For positive target amplitude $a(t)$, all neighbors with positive target overlap will also have positive scores. These neighboring curves can thus arise from a single represented target. Equation~\eqref{eq:overlap} supplies a geometric baseline for assessing apparent coactivation in the plots.

For a mean state $\bar{\mathbf x}$, an amplitude-matched target prediction uses $\hat a=\mathbf v_k^{\mathsf T}\bar{\mathbf x}$ and residual $\mathbf r=\bar{\mathbf x}-\hat a\mathbf v_k$. A nonzero residual can reflect decoding noise, dynamics, or other representational components; it is not by itself proof of competitor retrieval. Selection outputs $z_i$ and spike recordings provide complementary evidence. Moreover, with 613 candidate vectors in 512 dimensions, arbitrary lexical mixture coefficients cannot in general be recovered uniquely from the state vector alone.

%% file: tables/baseline.tex
\begin{tabular}{lrrrr}
\toprule
& \multicolumn{2}{c}{Production} & \multicolumn{2}{c}{Perception}\\
Word & Target score & Margin & Target score & Margin\\
\midrule
Tasse & 0.9541 & 0.4504 & 0.9540 & 0.5160\\
Tasche & 0.9726 & 0.4286 & 0.9729 & 0.4621\\
Kanne & 0.9729 & 0.4110 & 0.9649 & 0.4389\\
Katze & 0.9706 & 0.5624 & 0.9647 & 0.4550\\
Teller & 0.9621 & 0.5739 & 0.9792 & 0.4791\\
Topf & 0.9486 & 0.5791 & 0.9574 & 0.4372\\
Stuhl & 0.9464 & 0.4605 & 0.9481 & 0.6304\\
Schrank & 0.9616 & 0.4447 & 0.9533 & 0.5658\\
\bottomrule
\end{tabular}

%% file: sections/07-discussion.tex
\section{Discussion}
\subsection{What the Model Explained}
The model demonstrates how a structured lexical inventory can be coupled to neural retrieval circuits in two processing directions. Its explanatory value lies in making the relation between specification and behavior explicit. Shared graph components influence vector overlap; vector overlap determines match values; the associative-memory variants WTA selection (competition) and thresholded transfer regulate retrieval; and spiking populations approximate the resulting computations over time. The baseline results illustrate this chain for complete, unambiguous input vectors.

A particular strength of the model is that its internal processing can be inspected across all three lexical levels and the associative memories connecting them. The time courses in Section~\ref{sec:activation-interpretation} illustrate how decoded population activity in the conceptual, lemma, and phonological buffers can be examined alongside the selection outputs of the retrieval circuits. This access makes it possible to follow the development of a lexical representation through the processing pathway and to distinguish overlap with related representations from the selection of an additional lexical association. The ability to observe and interpret these internal activity patterns is itself an important outcome of the modeling work: it provides a basis for investigating how representational structure and neural retrieval dynamics jointly produce the observed output.

\subsection{Representational assumptions and emergent behavior}
Similarity is partly built into the model through its relational graphs. It would therefore be circular to treat elevated conceptual or phonological overlap alone as independent confirmation that the model has discovered linguistic similarity. The neural consequences of that structure are a different question. Whether a neighboring match crosses a threshold, survives competition, or alters an output trajectory depends on the retrieval circuit and its dynamics.

The connection between \emph{Tasche} and a vending-machine-related visual node illustrates how individual source associations can influence the resulting similarity structure. Assessing the linguistic adequacy of such associations requires comparison with human semantic judgments. Keeping these associations explicit in the documentation makes their effects traceable and provides a basis for testing alternative lexical specifications. Any revisions can then be recorded and evaluated separately from the behavior of the neural retrieval circuits.

The distinction also applies to the term \emph{coactivation}. Representational overlap, simultaneous selection outputs, and concurrent neuronal firing are different observations. A single distributed representation normally involves many active neurons, and a single selected word can yield several positive vocabulary projections. The geometric baseline in Eq.~\eqref{eq:overlap} makes this distinction testable. It prevents interpretive claims about lexical competition from being based solely on a non-target projection.

\subsection{Relation to earlier work}
The present work builds on the lexical modeling approach investigated in earlier studies of naming, cueing, and production--perception interactions \citep{stille2020,kroeger2016}. The motivation for this methodological account is to make the relationship between lexical data, vector construction, associative-memory operations, and measured neural activity explicit and inspectable. In particular, the paper specifies how within-level similarity arises from the encoded relations, how this differs from cross-level association, and how both affect the interpretation of activation curves. Documentation of the 613-entry inventory and access to the implementation and simulation resources support examination and reuse of these mechanisms. This contribution complements the task-oriented investigations in the earlier publications by providing a formal account of the current lexical core and a basis for subsequent controlled analyses.

The earlier naming study addressed lexical access and cueing within a broader task model \citep{stille2020}. The present account specifies the same basic lexical organization, with conceptual, lemma, and phonological representations linked by associative retrieval. Its narrower scope concerns the tasks used to investigate this organization and the surrounding task architecture; the full 613-entry lexical inventory remains available. The explanatory relationships between representational structure, associative retrieval, and neural activity concern this shared lexical organization. The present study does not claim to reproduce the earlier study's experimental conditions or findings. It also omits the recurrent monitoring architecture considered in earlier production--perception work \citep{kroeger2016}. The current direct concept buffer replaces the additional conceptual-access and cognition stages present in the original WWT implementation developed by Katharina Stille.

The change in random-vector generation, the use of 512 dimensions in the present configuration, and the different surrounding task architecture make the simulations of \citet{stille2020} and the current simulations distinct model realizations. Preserving a recursive construction rule is not equivalent to preserving all vectors, neural weights, or time courses. This distinction permits the present software to serve as a transparent implementation without overstating replication.

\subsection{Comparison with other lexical-access architectures}
\label{sec:lexical-architectures}
The distinction between selecting a lexical identity and retrieving its phonological form connects the present production pathway to established accounts of word production. In WEAVER++, lemma selection precedes word-form encoding, and the selected lemma provides access to its form \citep{levelt1999}. Competition is expressed through a selection rule sensitive to a candidate's activation relative to that of the alternatives. Our production pathway implements a comparable functional division through WTA selection (competition) from concept to lemma, followed by thresholded transfer from lemma to phonological form. The mechanisms differ: WEAVER++ uses a probabilistic selection rule with verification of the relevant lexical links, whereas our first associative memory uses lateral inhibition between neural candidate populations. Moreover, our memories remain continuously connected, so downstream activity follows the evolving lemma state, including selection transients. The timing of these successive transformations therefore emerges from the ongoing neural dynamics.

The interactive two-step model of \citet{dell1997} also separates word selection from phonological retrieval. Its semantic, word, and phoneme units exchange activation through bidirectional excitatory connections. Word selection identifies the most active eligible word, which then receives an activation boost before phoneme selection. This organization allows phonological activation to influence word-level processing during retrieval. Our current implementation places feedback within the WTA selection (competition) circuit, through lateral inhibition, while connections between lexical levels follow the direction of the task. Conceptual and phonological structure enters through precomputed vector geometry, and the output is a complete phonological form vector in production or a concept vector in perception. These choices isolate the contribution of structured representations and associative retrieval; the interactive architecture additionally addresses reciprocal influences among lexical levels and the selection of individual phonemes.

A simpler mathematical reference is the linear heteroassociative map $\mathbf y=VU^{\mathsf T}\mathbf x$, which weights every output vector directly by its input match. Our memories insert the nonlinear selection mechanism $\mathcal F$ between these operations, as shown in Eq.~\eqref{eq:am}. This additional step supplies the thresholding and competition through which the model regulates which lexical associations contribute to the output.

These comparisons distinguish the selection rule, the direction of interlevel coupling, and the unit being retrieved as separate architectural choices. Within the present framework, using the thresholded transfer variant at both transitions would allow every sufficiently strong match to pass at either stage; using the WTA selection (competition) variant at both transitions would add a second competitive selection after lemma retrieval. Adding interlevel feedback would introduce another mechanism, allowing an emerging output representation to influence earlier processing. Such variants could be studied with the same lexicon and input conditions to examine how these choices affect selection dynamics and retrieval under ambiguous input. The current combination assigns a clear role to each memory: WTA selection (competition) of a lemma identity followed by thresholded transfer to its associated target representation.

\subsection{Neural mechanisms and probabilistic interpretation}
\label{sec:neural-probabilistic}
Neural and probabilistic descriptions address complementary aspects of a model. A neural description specifies how neurons and synapses represent information and transform it over time. A probabilistic description specifies probabilities over possible stimuli, interpretations, or responses and the rules for computing or updating them. Both descriptions can apply to the same system: neural population codes can represent probability distributions and support Bayesian inference \citep{ma2006}. Within the NEF, \citet{eliasmithmartens2011} demonstrate probabilistic inference and normalization in a recurrent spiking network. A model can therefore be neural in its realization and probabilistic in the computation it performs.

The present model specifies the dynamics of lexical representations, candidate selection, and associative transfer. Its plotted values have meanings defined by these operations. For example, a thresholded transfer output of 0.8 assigns weight 0.8 to its associated stored vector in Eq.~\eqref{eq:am}. An output near 1 in the WTA selection (competition) memory reflects the configured amplitude of a selected response. Likewise, a positive buffer projection onto \emph{Tasse} during retrieval of \emph{Kanne} can reflect shared vector components, as explained in Section~\ref{sec:activation-interpretation}. Interpreting a value of 0.8 as an 80\% probability that a particular word is intended would introduce an additional assumption about the meaning of that activity.

Such a probabilistic interpretation would start by defining the relevant alternatives and evidence. For lexical recognition, one possible quantity is the probability of each word given an ambiguous input and its context. A model would then specify how lexical expectations and input reliability determine that distribution and how neural activity encodes it. Rescaling activity values to sum to 1 establishes normalization; their interpretation as probabilities additionally depends on these modeling assumptions and their evaluation. The probabilistic selection rule in WEAVER++ discussed in Section~\ref{sec:lexical-architectures} concerns the chance of selecting a candidate \citep{levelt1999}; a probability that a word generated an input concerns lexical uncertainty. These examples illustrate why the meaning of a probability must be specified alongside its neural or algorithmic implementation. The present account focuses on the mechanisms that retrieve and transfer specified lexical representations.

\subsection{Limitations and next analyses}
\label{sec:limitations}
The scope of the present study and the next steps concern several distinct aspects of the model and its evaluation. We address these separately below.

\begin{enumerate}
\item \textbf{Lexical specification and representational scope.} The lexicon is manually specified and includes inherited associations and documented repairs. No lexical learning process is simulated. The lemma layer currently supplies identities without shared grammatical features. Phonological composition does not by itself establish sequence-sensitive encoding. The present tasks isolate lexical retrieval between specified representations; Section~\ref{sec:sensorimotor-interfaces} describes their extension toward acoustic input, articulation, and sound generation.

\item \textbf{Recurrent input maintenance.} A recurrent input buffer is a concrete candidate for the next implementation step (Section~\ref{sec:neural-implementation}). It would add short-term retention to the current input-driven architecture and permit retrieval to be evaluated after a brief cue has disappeared. Implementing this extension would involve specifying how a representation is loaded, held, and cleared or replaced, together with the recurrent gain and synaptic time constant. Evaluation would then examine the stability of the retained vector and its amplitude, as well as retrieval success across different cue durations and retention intervals. Keeping the lexical vectors and associative mappings fixed would allow the contribution of this maintenance mechanism to be assessed against the present sustained-input baseline.

\item \textbf{Coverage of the simulation corpus.} The simulation corpus defined in Section~\ref{sec:simulation-corpus} supports the paper's focus on model description and illustrative retrieval behavior. All 613 lexical entries participate as candidates in the associative memories, and each retrieval decision is evaluated against the full inventory. Input coverage comprises the eight selected words in each direction. This provides sixteen documented examples from the complete implemented lexicon. Extending the corpus with every lexical entry as input in both directions would provide a systematic assessment of retrieval across the inventory.

\item \textbf{Robustness across model configurations.} The reported simulations also use a fixed model configuration: the same parameter settings, vector seed, and network seed are retained across the selected words. This permits a reproducible comparison without item-specific retuning and establishes that one common configuration supports both tasks for the tested inputs. It does not yet quantify sensitivity to alternative random vectors, neural realizations, or parameter choices. The two sources of variation introduced in Sections~\ref{sec:graph-encoding} and~\ref{sec:neural-implementation} can be examined separately: repeating the benchmark across vector seeds would assess dependence on representational geometry, while varying network seeds with the vocabulary fixed would assess variability in its neural implementation. These tests concern robustness beyond the reported realization, rather than the validity of the observed retrieval results themselves.

\item \textbf{Relationship to human behavior.} The relationship to human behavior should likewise be assessed at distinct levels. The separation of concepts, lemmas, and phonological forms, together with the two processing directions, gives the model a psycholinguistically interpretable organization. Earlier work on naming and cueing provides a relevant behavioral context \citep{stille2020}, and the present specification makes the proposed mechanisms available for more detailed examination. Successful retrieval of the selected words nevertheless does not by itself establish quantitative agreement with human reaction times, error distributions, or similarity effects. Such agreement requires explicit comparisons with behavioral observations under corresponding task conditions. The absence of a new behavioral comparison in the present benchmark does not imply that the architecture is unrelated to human processing; it identifies a further step needed to assess the explanatory reach of this particular implementation. Findings from earlier models should inform that assessment while remaining distinguished from validation of the current implementation under the tasks examined here.

\item \textbf{Neuron ablation.} The simulation software also supports neuron ablation in each of the three state buffers and both associative memories of each processing pathway. A selected proportion of neurons can be silenced independently in each component, allowing the effects of reduced neural resources on representation and retrieval to be explored. The \href{https://www.speechtrainer.eu/nef-spa-lexical-model.htm}{online simulation page} provides selected recorded ablation examples; the local simulation supports freely adjustable ablation settings. The present paper documents the intact baseline, while a detailed treatment and systematic evaluation of ablation conditions are reserved for subsequent publications. Earlier insights into such effects are provided by \citet{stille2020}, who examined neuron ablation within lexical buffers and between-level associative memories in a naming task, including the effects of semantic and phonological cues.

\item \textbf{Controlled variation of representational overlap.} A useful next analysis would vary representational overlap while keeping the candidate inventory and neural allocation fixed, and then separate changes in state projections from changes in selection outputs. Repeated vector and network seeds would quantify two different sources of variability. Such experiments would strengthen the distinction between assumed representational structure and consequences of neural retrieval, but they are not needed to identify the mathematical meaning of the quantities documented here.
\end{enumerate}

\subsection{Connecting lexical processing to perception and articulation}
\label{sec:sensorimotor-interfaces}
The tasks in Section~\ref{sec:tasks} examine the lexical component of a broader speech-processing system. In a more complete perception task, an acoustic speech signal would enter an auditory processing pathway, which would construct the phonological representation used for lexical access. In production, the retrieved phonological form provides an intermediate representation on the way to an audible utterance. The broader task requires this form to initiate a temporally organized articulatory program, drive a vocal-tract model, and generate an acoustic output. These extensions would connect the present lexical transformations to observable speech perception and production.

A strength of the NEF-SPA approach is its capacity to integrate these cognitive, sensory, and motor processes. The SPA supplies a common representational format and operations for organizing higher-level processing, while also linking perceptual and motor representations to those computations \citep{eliasmith2013}. The NEF provides methods for realizing representations, transformations, and continuous dynamics in neural populations \citep{eliasmith2003}. Nengo supplies the implementation environment in which these neural populations can be connected to external sensory and motor components, including models of the body and its environment \citep{bekolay2014}. This combination supports the connection of lexical representations to models of auditory input and vocal-tract control within a common computational framework.

Spaun provides an established example of this integration \citep{eliasmith2012}. It receives images of handwritten or printed characters through a simulated eye and generates written responses by controlling a physically modeled arm. Its visual system supplies information for cognitive processing, and its motor system turns the resulting response into an observable movement. The model therefore demonstrates both sides of the connection: visual signals enter through a sensory interface, and answers emerge as writing through a motor effector. For the present speech model, the corresponding objective is to link auditory processing and lexical computation to articulatory movement and acoustic output.

Several components of this route are already realized in our current modeling program. The companion articulatory manuscript documents neural syllable timing, overlapping gesture activation, and the conversion of decoded control signals into vocal-tract movement \citep{kroegerArticulation2026}. An ordered syllable readout for the selected lexical entries provides a shared interface between the lexical and articulatory examples. The current components are evaluated in separate simulations, with symbolic program selection and replay at specified interfaces. A complementary acoustic component is provided by the DYNARTmo extension, which derives vocal-tract acoustics and aerodynamic behavior from articulatory control and geometry and generates speech signals \citep{kroegerAcoustics2026}. Together, these developments provide implemented components spanning lexical retrieval, articulatory control, and sound generation.

The next integration steps are to let accepted lexical outputs initiate the corresponding articulatory programs, connect the neural control stages in a common simulation, and pass their output to the acoustic model. On the perception side, developing an auditory front end that maps an incoming speech signal to the phonological representations required by the lexicon remains a further objective for the present implementation. The recurrent input buffer discussed in Section~\ref{sec:neural-implementation} could then support temporary maintenance of the resulting representation. These extensions would make it possible to evaluate lexical processing together with its sensory input and motor consequences, including the intelligibility of generated speech and the effects of variation in the acoustic input.

%% file: sections/08-resources.tex
\section{Reproducibility and Resource Availability}
The model repository is \href{https://github.com/bjkroeger/nef-spa-mental-lexicon}{\texttt{bjkroeger/nef-spa-mental-lexicon} on GitHub}. Detailed simulation documentation is available at \href{https://www.speechtrainer.eu/nef-spa-lexical-model.htm}{the NEF-SPA Lexical Model page at speechtrainer.eu}. The documentation distinguishes the lexical graphs, state projections, and memory-entry outputs and provides access to demonstrations. The public version corresponding to this paper is identified by \href{https://github.com/bjkroeger/nef-spa-mental-lexicon/tree/0a0ec39be0143c4d444823ebe652b00eb38638a3}{commit \texttt{0a0ec39be014}} (5 November 2026).

ChatGPT Codex played a substantial role in cleaning up and redeveloping the NEF-SPA implementation provided with this paper. Guided by the author's specifications and the architecture described in \citet{stille2020}, it independently carried out substantial parts of the code development. The author directed the scientific modeling decisions and reviewed the resulting implementation.

The local configuration examined here uses Python 3.13, Nengo 4.1.0, and NengoSPA 2.0.0. Lexical data, extraction provenance, deterministic vector generation, and recorded simulation parameters are provided in the project. The manuscript's accompanying provenance manifest records SHA-256 hashes of the local files used to construct its tables and figures. The eight-entry fixture is retained separately from the active 613-entry dataset.

The model source files and numerical benchmark report were checked against this public commit. The lexical dataset has identical model content; the public copy replaces a machine-specific source-directory path with a relative path. The displayed perception and production recordings have identical numerical content, differing only in recorded execution time. The accompanying provenance manifest documents these correspondences and the public file locations. Figure-generation scripts and the neuron-count audit are supplied as ancillary material with this preprint. The article's license applies to the manuscript and its accompanying materials; it does not relicense the separately hosted model code or lexical data, whose reuse is governed by the repository's own rights and provenance information.

%% file: sections/09-conclusion.tex
\section{Conclusion}
A precise description of a neural mental lexicon requires separate definitions of representational structure, cross-level association, and measured activity. The present NEF-SPA implementation makes these distinctions explicit for conceptual, lemma, and phonological levels. Recursive graph encoding determines within-level geometry, while the two associative-memory variants, WTA selection (competition) and thresholded transfer, implement directed lexical transfer in spiking populations. Baseline simulations establish correct retrieval of eight selected words from the lexicon, supplied as complete input representations, in both directions against a 613-entry inventory. The resulting specification provides a basis for inspecting the model, reproducing its computations, and designing targeted tests of the relationship between similarity and retrieval.

%% file: sections/appendix.tex
\section{Notation and Implementation Correspondence}
\label{app:implementation}
\begin{table}[H]
\centering\small
\begin{tabular}{lp{10.3cm}}\toprule
Symbol & Meaning\\\midrule
$D,N$ & Vector dimensions and number of lexical candidates\\
$\mathbf v_i^{(L)}$ & Unit-length vocabulary vector for entry $i$ at level $L$\\
$S^{(L)}_{ij}$ & Static within-level dot product of two vocabulary vectors\\
$U,V$ & Input and associated output vocabulary matrices\\
$q_i,z_i$ & Input match and decoded selection output for entry $i$\\
$\hat{\mathbf x}^{(L)}$ & Decoded time-dependent state vector\\
$A_i^{(L)}$ & Projection of the decoded state onto a vocabulary vector\\\bottomrule
\end{tabular}
\end{table}
The dataset's \texttt{entries} array defines lexical correspondences through the fields \texttt{word}, \texttt{concept}, \texttt{lemma}, and \texttt{phonology}, together with an index into the original dataset developed by Katharina Stille. The implementation key \texttt{phonology} denotes the phonological level. The \path{concept_network} and \path{phonology_network} objects map node identifiers to lists of relation--target pairs. The \path{phonology_repairs} records document modifications needed to construct the active inventory. Source checksums identify the original input files used by Katharina Stille.

The implementation of Eq.~\eqref{eq:recursive} is in \texttt{lexicon\_vectors.py}. Directed networks are defined in \texttt{perception.py} and \texttt{production.py}; the full-candidate evaluation is implemented in \path{benchmark_full_lexicon.py}. This evaluation script runs the eight selected words in each direction and writes \path{baseline_benchmark.json}, the numerical report underlying Table~\ref{tab:baseline}. The report records the model settings and, for each of the sixteen trials, the input, retrieved entry, target score, margin, and highest-ranked candidates. It is an output of the model evaluation.

The saved files \path{perception-Kanne.json} and \path{production-Kanne.json} provide the activity traces used in Figure~\ref{fig:traces}. In the public repository, the numerical report is stored at \path{docs/baseline_benchmark.json}, and the recordings are stored under \path{website/deploy/perception-production-model/runs-613-20261104/}. The local analysis uses copies under \path{results/lexicon613/}; the ancillary provenance manifest records the correspondence. Rounded, downsampled traces are used for displaying the time courses, while the retrieval table uses the scores retained in the numerical evaluation report. The separate figure-generation script constructs the static similarity matrices from the lexical data and vector seed and reads these saved outputs to prepare the figures and table. The accompanying provenance manifest identifies the source and result files by their checksums.